\documentclass[runningheads]{llncs}
\usepackage[T1]{fontenc}
\usepackage{graphicx,verbatim}
\usepackage{amsmath,amssymb}
\usepackage{multirow}
\usepackage{pifont}
\usepackage{xcolor}

\newcommand{\cmark}{\ding{51}} 
\newcommand{\xmark}{\ding{55}} 
\begin{document}
\title{Seeing Through Smoke: Surgical Desmoking for Improved Visual Perception} 
%
\author{Jingpei Lu \orcidID{0000-0002-9136-6096} \and
Fengyi Jiang \orcidID{0009-0007-9420-5259} \and
Xiaorui Zhang \orcidID{0009-0005-3639-3875} \and
Lingbo Jin \and
Omid Mohareri
}
\authorrunning{J. Lu et al.}
%
\institute{Intuitive Surgical, Sunnyvale, CA 94086, USA\\
\email{jingpei.lu@intusurg.com}\\
}



\maketitle              
\begin{abstract}

Minimally invasive and robot-assisted surgery relies heavily on endoscopic imaging, yet surgical smoke produced by electrocautery and vessel-sealing instruments can severely degrade visual perception and hinder vision-based functionalities. We present a transformer-based surgical desmoking model with a physics-inspired desmoking head that jointly predicts smoke-free image and corresponding smoke map. To address the scarcity of paired smoky-to-smoke-free training data, we develop a synthetic data generation pipeline that blends artificial smoke patterns with real endoscopic images, yielding over 80,000 paired samples for supervised training.
We further curate, to our knowledge, the largest paired surgical smoke dataset to date, comprising 5,817 image pairs captured with the da Vinci robotic surgical system, enabling benchmarking on high-resolution endoscopic images. Extensive experiments on both a public benchmark and our dataset demonstrate state-of-the-art performance in image reconstruction compared to existing dehazing and desmoking approaches. We also assess the impact of desmoking on downstream stereo depth estimation and instrument segmentation, highlighting both the potential benefits and current limitations of digital smoke removal methods. 


\end{abstract}
\section{Introduction}

Minimally invasive procedures, especially robot-assisted surgery, can improve patient outcomes by reducing incision size, infection risk, and recovery time versus open surgery. However, they introduce visibility challenges when high-energy instruments such as electrocautery and vessel sealers are used. Rapid tissue heating vaporizes intracellular water and degrades organic components, producing surgical smoke made of water vapor and protein/lipid byproducts. Because minimally invasive surgery relies heavily on endoscopic imaging, smoke and lens fog can obscure fine tissue detail, prolong operative time, and increase error risk by limiting informed intraoperative decision-making. Smoke artifacts also degrade downstream computer-assisted functions, including instrument segmentation, tool tracking, and 3D reconstruction \cite{ding2024segstrong}.

Common smoke evacuation methods include instrument-tip suction, port-side evacuation, and valveless trocar systems, each with limitations \cite{zhang2025methods}. For example, valveless insufflation systems such as AirSeal can draw room air into the abdomen during suction, highlighting trade-offs among smoke clearance, pressure control, and gas composition \cite{weenink2020airseal}. Alternatively, digital smoke removal is appealing because it can enhance visibility in real time without interrupting the procedure, reducing the need for manual lens cleaning or frequent smoke-evacuation maneuvers. It may also reduce reliance on specialized hardware, lowering device cost and increasing surgeon autonomy. Nevertheless, desmoking is challenging because smoke varies widely in density and motion and induces non-uniform scattering that can resemble anatomical texture.

Early analytical methods use Bayesian probabilistic graphical models \cite{baid2017joint,kotwal2016joint} or dark channel priors \cite{he2010single,tchaka2017chromaticity} to suppress smoke and enhance contrast. However, these approaches often rely on strong priors developed for natural hazy images, overlooking key differences between surgical smoke and atmospheric haze. More recently, domain-specific, data-driven methods have been proposed for desmoking in minimally invasive surgery. Several approaches apply convolutional neural networks (CNNs) \cite{bolkar2018deep,kanakatte2021surgical,wang2019multiscale,DarkChannelCNN} to produce smoke-free images in real time without explicit physical modeling. Further gains in color fidelity and contrast have been achieved using attention-based networks \cite{hong2023mars,GenerativeSmoke,ffanet}, hybrid CNN--transformer architectures \cite{wang2023surgical}, and diffusion models with multi-frequency learning \cite{li2024multi}.
Progress on surgical desmoking remains limited by the scarcity of paired \emph{in vivo} data with smoke-free ground truth. Although unpaired learning can reduce the need for paired ground truth \cite{selfSVD,pan2022desmoke}, it often introduces structural distortions or color shifts that reduce clinical reliability. A practical alternative is to train on synthetic data while explicitly addressing the sim-to-real domain gap.

In this work, we propose a transformer-based model for surgical desmoking. Our architecture comprises a Vision Transformer (ViT) \cite{vit} backbone, a DPT decoder \cite{dpt}, and a physics-inspired desmoking head that predicts a smoke-free image along with a smoke map.
To address data scarcity, we generate a large synthetic training set by blending artificial smoke patterns into real endoscopic images. For evaluation, we also collect 5,817 paired smoky-to-smoke-free images acquired with Intuitive Surgical's da Vinci robotic surgical system.
We evaluate our method on our collected dataset and on the \emph{in vivo} benchmark dataset \cite{de-smoking}, and demonstrate state-of-the-art performance.
Beyond image reconstruction, we further study the effect of desmoking on downstream stereo depth estimation and instrument segmentation. 
The results indicate that our method can potentially provide benefits to downstream computer vision algorithms, representing an important step toward computer-assisted robotic surgery.


\section{Methodology}

In this paper, we introduce an end-to-end model for surgical desmoking (Section~\ref{sec:neuralnetwork}). To train the network, we develop a synthetic smoke data generation pipeline that produces large-scale paired smoky/smoke-free images (Section~\ref{sec:syntheticdata}). We also collect, to our knowledge, the largest paired surgical smoke dataset to benchmark desmoking performance (Section~\ref{sec:data_collection}). The dataset comprises 5,817 image pairs acquired in an \emph{ex vivo} setup and includes depth maps and segmentation labels to enable extensive downstream evaluation.

\subsection{Surgical Desmoking Model}
\label{sec:neuralnetwork}

\begin{figure}[t]
  \centering
  \includegraphics[width=0.9\linewidth]{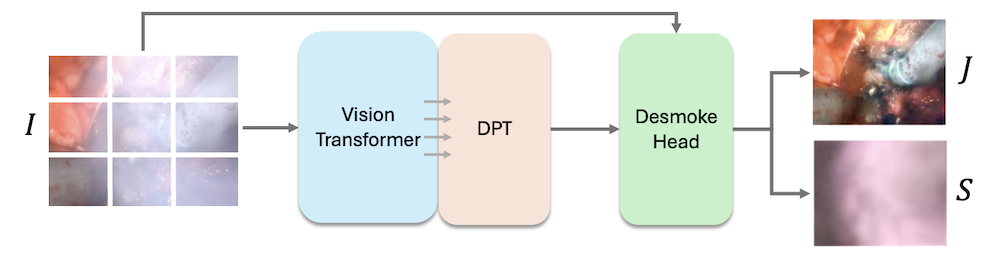}
  \caption{Overview of the proposed surgical desmoking model.}
  \label{fig:network}
\end{figure}

Figure~\ref{fig:network} provides an overview of the proposed desmoking model. Given an input smoky image $I \in \mathbb{R}^{H\times W \times 3}$, we use a ViT-Base backbone to extract multi-scale features, initialized with Masked Autoencoder pretrained weights \cite{he2022masked} on a large corpus of surgical images. We fuse latent representations from intermediate ViT blocks $\{3,6,9,12\}$ and decode them using DPT blocks. Finally, we attach a lightweight desmoking head to jointly predict the smoke-free image $J$ and the corresponding smoke map $S$.

To design the desmoking head, we draw inspiration from single-image dehazing methods~\cite{he2010single,song2023vision}. We assume a standard atmospheric scattering model~\cite{mccartney1976optics}:
\begin{equation}
  I = Jt + A(1-t),
\end{equation}
where $I$ denotes the input smoky image, $J$ the smoke-free image, $A$ the global atmospheric light, and $t$ the medium transmission map. Our objective is to recover $J$ via:
\begin{equation}
  J = \frac{I - A + At}{t}.
\end{equation}
Following~\cite{song2023vision}, we can rewrite Eq.~(2) as:
\begin{equation}
  J = KI - B + I,
\end{equation}
where $K = 1/t -1$ and $B = A(1/t -1)$. We design the desmoke head to directly predict $K$ and $B$ by regressing a 4D image $O = [K,B] \in \mathbb{R}^{H\times W \times 4}$, such that the network architecture softly constrains the relationship between $K$ and $B$.

Since $I$ can be viewed as a combination of the attenuated scene radiance $Jt$ and the airlight term $A(1-t)$ (light scattered by particles), the smoke map is naturally correlated with the airlight component. We therefore predict the smoke map $S$ as a function of $A(1-t)$:
\begin{equation}
  S = f(A(1-t)) = f\!\left(\frac{B}{K+1}\right),
\end{equation}
where $f(\cdot)$ is approximated as a convolutional layer.

\subsection{Training and Synthetic Data Generation}
\label{sec:syntheticdata}

\begin{figure}[t]
  \centering
  \includegraphics[width=0.9\linewidth]{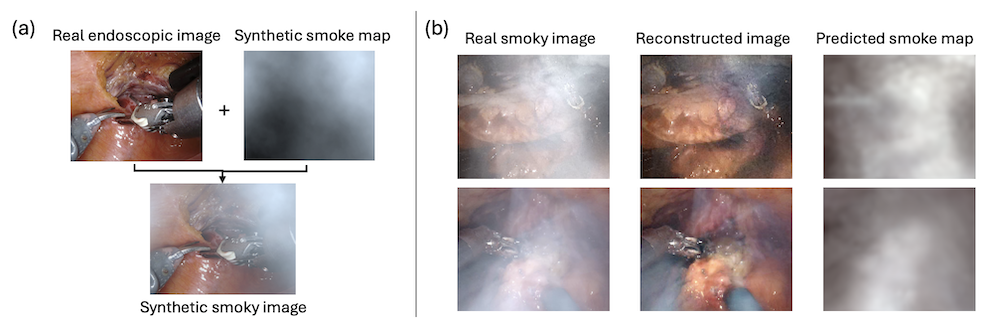}
  \caption{(a) Synthetic smoke data generation pipeline. (b) Qualitative results for desmoking and smoke-map prediction on real smoky images.}
  \label{fig:synthetic_data}
\end{figure}

Collecting large-scale surgical smoke data is labor-intensive and costly, and obtaining paired smoky-to-smoke-free images is even more challenging. Mining such paired samples from existing laparoscopic datasets is only feasible at limited scale \cite{lsd3k,selfSVD}. Consequently, synthetic smoke generation remains the most efficient approach for producing large-scale paired training data.

We curated a dataset of 10k smoke-free endoscopic images spanning both \emph{in vivo} and \emph{ex vivo} scenes. We then generated a diverse set of smoke maps with varying densities and spatial patterns. Finally, we synthesized smoky images by blending smoke maps with real endoscopic images, as illustrated in Fig.~\ref{fig:synthetic_data}(a). We use alpha blending with RGB channel correction and illumination enhancement to improve visual realism. To mitigate the sim-to-real gap, we apply domain randomization over smoke color, transparency, and blending coefficients. In total, we generate approximately 80k paired smoky/clean images at $512\times640$ resolution to train the proposed surgical desmoking network.

During training, we use $L_1$ (mean absolute error) losses for both smoke-free image reconstruction and smoke-map prediction:
\begin{equation}
  \mathcal{L} = \frac{1}{N} \sum_{i=1}^{N} \left(\lvert J^{gt}_i - J_i \rvert + \lambda \lvert S^{gt}_i - S_i \rvert\right),
\end{equation}
where $J^{gt}$ and $S^{gt}$ denote the ground-truth smoke-free image and smoke map, respectively, and we set $\lambda = 0.1$.
We train the network for 300 epochs using AdamW with a cosine-annealed learning rate from $10^{-5}$ to $10^{-7}$.
Because ground-truth smoke maps are difficult to obtain for real smoky scenes, we report only qualitative results for these cases in Fig.~\ref{fig:synthetic_data}(b). Although we do not evaluate smoke-map estimation quantitatively, we view it as an auxiliary output with potential future uses, such as confidence-aware smoke evacuation control, or guiding downstream tasks by masking high-smoke regions.

\subsection{Benchmark Smoke Data Acquisition}
\label{sec:data_collection}

To validate desmoking performance on experimental scenes, we collected data using an Intuitive Surgical da~Vinci robotic surgical system. We placed porcine tissue specimens (e.g., liver, intestine, stomach, and kidney) in a sealed chamber and used a smoke machine (LENSGO Portable Fog Machine) to generate smoke. At the beginning of each capture, the endoscope was positioned at a fixed viewpoint toward the specimens, and instrument manipulation was introduced. The smoke machine was then activated to introduce smoke into the chamber. After a short interval, we deactivated the smoke machine and allowed the smoke to distribute throughout the scene. Finally, a suction apparatus was activated to gradually evacuate the smoke.

We employed a bidirectional capture strategy---progressive occlusion followed by progressive clearance---which yields paired observations of identical static scenes across a continuous spectrum of smoke densities. To the best of our knowledge, the resulting dataset is the largest publicly available endoscopic smoke dataset to date (Table~\ref{tab:dataset}), addressing a critical gap in the availability of large-scale, high-quality paired data for surgical smoke removal. We further derived depth maps and tool segmentation labels from the smoke-free images to support downstream task evaluation and future investigations in surgical desmoking.

\begin{table}[t]
  \centering
  \caption{Existing surgical smoke datasets.}
  \label{tab:dataset}
  \setlength{\tabcolsep}{4pt}
  \begin{tabular}{l|c|c|c|c}
    \hline
    Dataset & Smoke type & Size & Image type (resolution) & Availability \\
    \hline
    LSD3K~\cite{lsd3k} & Synthetic & 3{,}000 pairs & Mono (480$\times$480) & \xmark \\
    LSVD~\cite{selfSVD} & Real & 786 clips & Mono (1080$\times$1920) & \xmark \\
    De-Smoking~\cite{de-smoking} & Real & 3{,}464 pairs & Mono (inconsistent) & \cmark \\
    Ours & Real & 5{,}817 pairs & Stereo (1024$\times$1280) & \cmark \\
    \hline
  \end{tabular}
\end{table}

\section{Experiments and Results}

We selected eight representative methods from the dehazing and desmoking literature for comparison, based on their reported performance and the availability of open-source implementations. We reimplemented the algorithms in~\cite{he2010single,song2023vision,ffanet,DarkChannelCNN,DualResidualNet,GenerativeSmoke,StructureReprNet,selfSVD} following the authors' released code, and evaluated them on the De-smoking dataset~\cite{de-smoking} as well as our in-house benchmark dataset.\sloppy
We found that most methods failed to produce reasonable results at $1024\times 1280$ resolution. Therefore, we resized all collected images to $512\times 640$ for the evaluations in this section. Our model has a size of 111.36~MB and runs at 52~FPS on an NVIDIA RTX~6000 GPU at $512\times 640$ resolution. We split our benchmark data into a fine-tuning set (3,097 image pairs) and a test set (2,720 image pairs) with disjoint anatomical structures. All numbers reported in Tables~3 are computed on the test set. We use the fine-tuning set to further fine-tune our model for an additional 50 epochs; these results are reported as \emph{Ours-finetune}. The model without fine-tuning on real smoke data is reported as \emph{Ours}.

\subsection{Surgical Desmoking Benchmark}

\begin{figure}[t]
  \centering
  \includegraphics[width=0.9\linewidth]{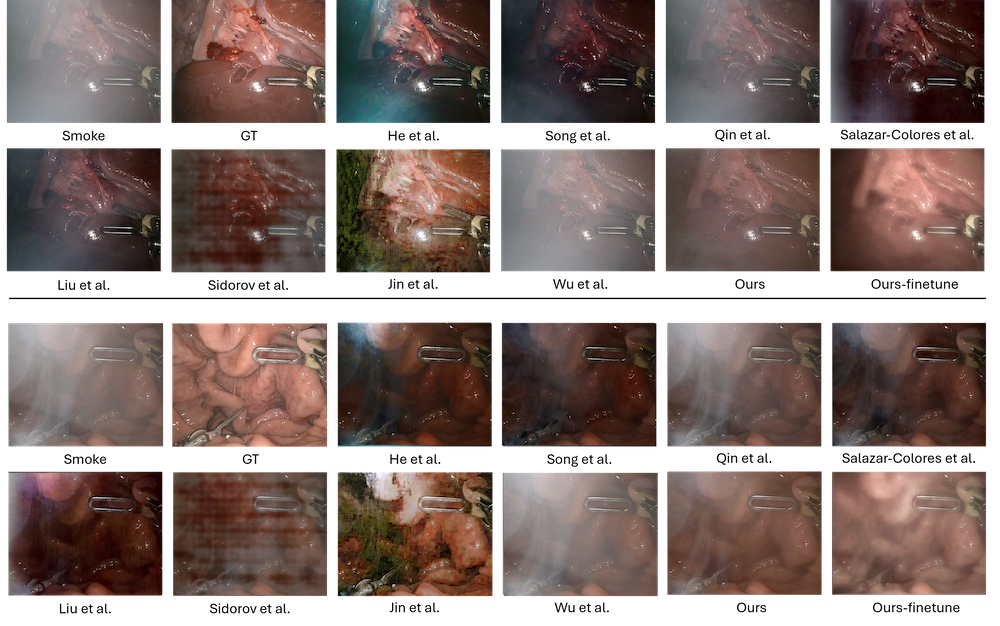}
  \caption{Qualitative comparisons on our test set across representative scenes (top: liver; bottom: stomach)}
  \label{fig:test}
\end{figure}

\begin{table}[t]
  \centering
  \caption{Quantitative results on the De-Smoking dataset}\label{tab:desmoke}
  \begin{tabular}{l|c|c|c|c}
    \hline
    \multirow{2}{*}{Method} & \multicolumn{2}{c|}{Cholecystectomy} & \multicolumn{2}{c}{Prostatectomy} \\
    \cline{2-5}
    & SSIM & PSNR & SSIM & PSNR\\
    \hline
    He et al.~\cite{he2010single} & 0.73 $\pm$ 0.09  & 18.44 $\pm$ 3.03 &  0.72 $\pm$ 0.12 & 19.39 $\pm$ 3.38\\
    Song et al.~\cite{song2023vision} & 0.83 $\pm$ 0.10 & 23.25 $\pm$ 4.51 & 0.73 $\pm$ 0.15 & 19.22 $\pm$ 5.93 \\
    Qin et al.~\cite{ffanet} &  0.84 $\pm$ 0.10 & 24.49 $\pm$ 5.88 & 0.72 $\pm$ 0.15 & 19.13 $\pm$ 6.68 \\
    Salazar et al.~\cite{DarkChannelCNN} &  0.75 $\pm$ 0.12 & 20.06 $\pm$ 4.00 & 0.72 $\pm$ 0.13 & 19.94 $\pm$ 4.10\\
    Liu et al.~\cite{DualResidualNet} &  0.81 $\pm$ 0.09 & 22.21 $\pm$ 3.79 & 0.73 $\pm$ 0.13 & 20.08 $\pm$ 5.70\\
    Sidorov et al.~\cite{GenerativeSmoke} &  0.76 $\pm$ 0.09 & 20.69 $\pm$ 3.89 & 0.66 $\pm$ 0.11 & 20.09 $\pm$ 2.82 \\
    Jin et al.~\cite{StructureReprNet} &  0.74 $\pm$ 0.13 & 21.53 $\pm$ 3.89 & 0.62 $\pm$ 0.14 & 18.92 $\pm$ 3.79 \\
    Wu et al.~\cite{selfSVD} &  0.83 $\pm$ 0.11 & 23.88 $\pm$ 6.89 & 0.67 $\pm$ 0.15 & 17.99 $\pm$ 6.13\\
    Ours  & \textbf{0.86 $\pm$ 0.09}&  \textbf{25.51 $\pm$ 4.69} & \textbf{0.75 $\pm$ 0.13} & \textbf{21.71 $\pm$ 4.41}\\
    \hline
  \end{tabular}
\end{table}

Table~2 reports the performance of our desmoking model against eight selected dehazing/desmoking baselines on the De-Smoking dataset~\cite{de-smoking}. The dataset comprises 1,063 paired images from cholecystectomy and 2,401 paired images from prostatectomy procedures, curated from \emph{in vivo} video clips using motion-tracking techniques. We evaluate smoke-free image reconstruction using structural similarity (SSIM) and peak signal-to-noise ratio (PSNR).
The values in Table~2 may differ slightly from those reported in~\cite{de-smoking} due to implementation details.
As shown, our method consistently outperforms existing approaches on both the cholecystectomy and prostatectomy subsets.

Table~3 reports performance on our benchmark dataset, and Fig.~\ref{fig:test} provides qualitative comparisons, showing that our method can recover scene content behind dense smoke with minimal color distortion and fewer artifacts. Because our dataset is evaluated at higher resolution, several competing methods degrade noticeably relative to their performance on lower-resolution benchmarks. We report results for \emph{Ours-finetune} for completeness, but exclude it from direct comparison because the competing models were not fine-tuned on real smoke images. Nevertheless, our approach achieves the best overall performance, with an average SSIM of 0.73 and PSNR of 20.61.

\begin{figure}[t]
  \centering
  \includegraphics[width=0.9\linewidth]{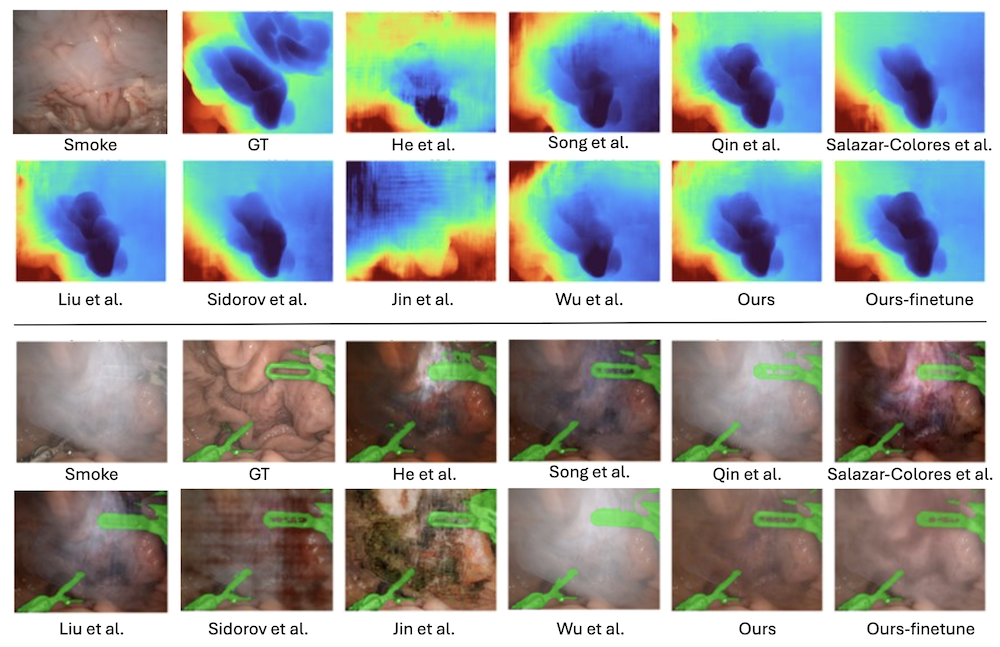}
  \caption{Qualitative results on downstream stereo depth estimation (top) and instrument segmentation (bottom) after different desmoking methods.}
  \label{fig:depth}
\end{figure}

\begin{table}[t]
  \centering
  \caption{Quantitative results on our test set for different desmoking methods. We report reconstruction quality (SSIM/PSNR) and downstream performance after desmoking: stereo depth estimation (MAE) and instrument segmentation (IoU).}\label{tab:ismoke_depth_seg}
  \begin{tabular}{l|c|c|c|c}
    \hline
    \multirow{2}{*}{Method} & \multicolumn{2}{c|}{Reconstruction} & Depth & Segmentation \\
    \cline{2-5}
    & SSIM $\uparrow$ & PSNR $\uparrow$ & MAE (mm) $\downarrow$ & $\%$ IoU $\uparrow$ \\
    \hline
    None & -- & -- & \textbf{47.22 $\pm$ 61.48} & 81.15 $\pm$ 17.01 \\
    He et al.~\cite{he2010single} &  0.53 $\pm$ 0.07 & 12.83 $\pm$ 1.53 & 51.53 $\pm$ 66.71 & 80.81 $\pm$ 17.03\\
    Song et al.~\cite{song2023vision} & 0.61 $\pm$ 0.09 & 14.45 $\pm$ 1.83 & 49.00 $\pm$ 66.45 & 80.90 $\pm$ 16.27\\
    Qin et al.~\cite{ffanet} & 0.72 $\pm$ 0.09  & 19.57 $\pm$ 3.20 & 49.03 $\pm$ 63.84 & 81.15 $\pm$ 17.15 \\
    Salazar et al.~\cite{DarkChannelCNN} &  0.54 $\pm$ 0.12 & 14.21 $\pm$ 2.05 &  53.73 $\pm$ 66.50 & 79.46 $\pm$ 16.85 \\
    Liu et al.~\cite{DualResidualNet} &  0.65 $\pm$ 0.11 & 16.43 $\pm$ 2.63 &  48.02 $\pm$ 63.62 & 81.31 $\pm$ 16.18 \\
    Sidorov et al.~\cite{GenerativeSmoke} &  0.59 $\pm$ 0.08 & 16.17 $\pm$ 1.86 &  59.44 $\pm$ 74.02 & 77.78 $\pm$ 16.70 \\
    Jin et al.~\cite{StructureReprNet} &  0.50 $\pm$ 0.15 & 17.41 $\pm$ 2.75 &  56.48 $\pm$ 69.13 & 74.98 $\pm$ 17.36 \\
    Wu et al.~\cite{selfSVD} &  0.70 $\pm$ 0.09 & 19.73 $\pm$ 4.26 &  53.03 $\pm$ 65.20 & 79.30 $\pm$ 17.13 \\
    Ours & \textbf{0.73 $\pm$ 0.09} & \textbf{20.61 $\pm$ 3.12} & 49.42 $\pm$ 64.48 & \textbf{81.48 $\pm$ 16.85} \\
    \hline
    Ours-finetune & 0.75 $\pm$ 0.09 & 23.57 $\pm$ 3.35 & 47.64 $\pm$ 62.03 & 81.81 $\pm$ 16.08 \\
    \hline
  \end{tabular}
\end{table}

\subsection{Impacts on Downstream Computer Vision Tasks}

Beyond image reconstruction, we investigate how surgical desmoking affects downstream computer vision tasks. We generate pseudo-ground-truth depth maps by running FoundationStereo~\cite{wen2025foundationstereo} on the smoke-free images, and obtain instrument segmentation labels using Segment Anything 3 (SAM3)~\cite{carion2025sam}. To assess the effect of desmoking, we first apply each method to the smoky images and then run FoundationStereo and SAM3 on the corresponding desmoked outputs. We evaluate depth estimation using mean absolute error (MAE) with respect to the pseudo-ground-truth depth, and evaluate segmentation using intersection-over-union (IoU) against the generated labels. Table~\ref{tab:ismoke_depth_seg} summarizes the quantitative results, while Fig.~\ref{fig:depth} provide qualitative comparisons.

Interestingly, for stereo depth estimation, the no-desmoking baseline (\emph{None}) performs best. One possible explanation is that foundation models such as FoundationStereo are relatively robust to smoke due to large-scale training.
In addition, desmoking can alter image statistics and potentially disrupt left--right feature consistency, which is critical for stereo matching, thereby degrading reconstruction accuracy. Overall, our results suggest that current desmoking approaches do not consistently improve stereo depth estimation under smoke.
In contrast, although some methods reduce segmentation performance, our approach improves IoU on smoky scenes. We attribute this to enhanced local contrast and clearer instrument boundaries after desmoking. While these findings reveal limitations of using desmoking purely as a preprocessing step, they also highlight its potential to benefit downstream applications (e.g., instrument segmentation) by improving visual perception in the presence of smoke.


\section{Conclusion}

We presented a transformer-based surgical desmoking framework that combines a ViT backbone with a physics-inspired head to jointly predict smoke-free images and corresponding smoke maps. To mitigate the scarcity of paired training data, we developed a large-scale synthetic data generation pipeline and curated a high-resolution paired dataset for benchmarking, enabling supervised training and evaluation at unprecedented scale.
Our method achieves state-of-the-art reconstruction performance across datasets, improving anatomical visibility under dense smoke while preserving visual fidelity.
Analysis of downstream tasks shows that desmoking can improve instrument segmentation, but does not consistently benefit stereo depth estimation, underscoring the nuanced interaction between image reconstruction and geometry-based algorithms.
Overall, these results position digital desmoking as a valuable component of surgical vision pipelines; future work will focus on real-time deployment and tighter integration with computer-assisted interventions.

%
%
%
\bibliographystyle{splncs04}
\bibliography{mybibliography}
\end{document}